%% file: main.tex
\documentclass{article} %

\PassOptionsToPackage{hyphens}{url}

\usepackage{iclr2023_conference,times}

\input{math_commands.tex}

\usepackage{microtype}
\usepackage{amsmath}
\usepackage{booktabs}
\usepackage{colortbl}
\usepackage[utf8]{inputenc}
\definecolor{lightgray}{rgb}{0.9,0.9,0.9}
\usepackage{caption}
\usepackage{subcaption}
\usepackage{xcolor}
\usepackage{graphicx}
\usepackage{setspace}
\usepackage[hidelinks]{hyperref}
\usepackage{url}
\usepackage{multirow}
\usepackage{colortbl}
\usepackage{tabularx}
\usepackage{blindtext}
\usepackage{pgfplots}
\pgfplotsset{compat=1.18} 
\usepackage{tikz}
\usetikzlibrary{er,positioning,bayesnet}
\usepackage[inline]{enumitem}
\usepackage{makecell}
\usepackage{siunitx}
\usepackage{nicefrac}
\usepackage{listings}
\usepackage[raster,skins]{tcolorbox} %
\usepackage{adjustbox}

\title{Qwen2 Technical Report}
\author{
\\
\parbox{\linewidth}{An Yang, Baosong Yang, Binyuan Hui, Bo Zheng, Bowen Yu, Chang Zhou, Chengpeng Li, Chengyuan Li, Dayiheng Liu, Fei Huang, Guanting Dong, Haoran Wei, Huan Lin, Jialong Tang, Jialin Wang, Jian Yang, Jianhong Tu, Jianwei Zhang, Jianxin Ma, Jianxin Yang, Jin Xu, Jingren Zhou, Jinze Bai, Jinzheng He, Junyang Lin, Kai Dang, Keming Lu, Keqin Chen, Kexin Yang, Mei Li, Mingfeng Xue, Na Ni, Pei Zhang, Peng Wang, Ru Peng, Rui Men, Ruize Gao, Runji Lin, Shijie Wang, Shuai Bai, Sinan Tan, Tianhang Zhu, Tianhao Li, Tianyu Liu, Wenbin Ge, Xiaodong Deng, Xiaohuan Zhou, Xingzhang Ren, Xinyu Zhang, Xipin Wei, Xuancheng Ren, Xuejing Liu, Yang Fan, Yang Yao, Yichang Zhang, Yu Wan, Yunfei Chu, Yuqiong Liu, Zeyu Cui, Zhenru Zhang, Zhifang Guo, and Zhihao Fan}
\AND
Qwen Team, Alibaba Group\thanks{Authors are ordered alphabetically by the first name.}\\
}

\iclrfinalcopy
\begin{document}

\maketitle

\begin{abstract}
This report introduces the Qwen2 series, the latest addition to our large language models and large multimodal models. 
We release a comprehensive suite of foundational and instruction-tuned language models, encompassing a parameter range from 0.5 to 72 billion, featuring dense models and a Mixture-of-Experts model. 
Qwen2 surpasses most prior open-weight models, including its predecessor Qwen1.5, and exhibits competitive performance relative to proprietary models across diverse benchmarks on language understanding, generation, multilingual proficiency, coding, mathematics, and reasoning.

\smallskip
The flagship model, Qwen2-72B, showcases remarkable performance: 84.2 on MMLU, 37.9 on GPQA, 64.6 on HumanEval, 89.5 on GSM8K, and 82.4 on BBH as a base language model. 
The instruction-tuned variant, Qwen2-72B-Instruct, attains 9.1 on MT-Bench, 48.1 on Arena-Hard, and 35.7 on LiveCodeBench. 
Moreover, Qwen2 demonstrates robust multilingual capabilities, proficient in approximately 30 languages, spanning English, Chinese, Spanish, French, German, Arabic, Russian, Korean, Japanese, Thai, Vietnamese, and more, underscoring its versatility and global reach.

\smallskip
To foster community innovation and accessibility, we have made the Qwen2 model weights openly available on Hugging Face\footnote{\url{https://huggingface.co/Qwen}} and ModelScope\footnote{\url{https://modelscope.cn/organization/qwen}}, and the supplementary materials including example code on GitHub\footnote{\url{https://github.com/QwenLM/Qwen2}}. 
These platforms also include resources for quantization, fine-tuning, and deployment, facilitating a wide range of applications and research endeavors.

\end{abstract}
\clearpage

\tableofcontents
\clearpage

\input{content/intro.tex}

\input{content/model_tokenizer.tex}

\input{content/pretraining}

\input{content/posttraining}

\input{content/experiments.tex}
\input{content/conclusion}

\clearpage

\bibliography{biblio}
\bibliographystyle{iclr2023_conference}
\clearpage

\end{document}

%% file: math_commands.tex
\usepackage{amsmath,amsfonts,bm}

\def\eqref#1{equation~\ref{#1}}

\def\1{\bm{1}}

\DeclareMathAlphabet{\mathsfit}{\encodingdefault}{\sfdefault}{m}{sl}
\SetMathAlphabet{\mathsfit}{bold}{\encodingdefault}{\sfdefault}{bx}{n}

%% file: content/intro.tex
\section{Introduction}
\label{sec:intro}

Following the emergence of ChatGPT~\citep{chatgpt}, enthusiasm for large language models (LLMs) has escalated globally. 
The release of the Llama series~\citep{llama} has further ignited interests within the open-source community, particularly regarding GPT-level local LLMs.
Recently, Claude-3 Opus~\citep{claude3} and GPT-4o (omni)~\citep{gpt4o}, the updated model for ChatGPT, have ascended to the pinnacle of the Chatbot Arena~\citep{arena} in quick succession. This platform is well-regarded for its human evaluations of LLMs.
Moreover, Llama-3~\citep{llama3} has emerged as the state-of-the-art open-weight model series, narrowing the performance gap with leading proprietary models and widely acknowledged as GPT-4--level.
An increasing number of competitive LLMs are now pursuing advancements similar to those made by the GPT series from OpenAI. 
Many of these models, including Qwen~\citep{qwen}, Mistral~\citep{mistral}, 
Gemma~\citep{gemma}, etc., have been released in an open-weight manner.

Over recent months, we have successively introduced the Qwen series~\citep{qwen} and progressed to Qwen1.5~\citep{qwen1.5}.
In the meantime, we have unveiled the vision-language model Qwen-VL~\citep{qwenvl}, and launched the audio-language model Qwen-Audio~\citep{qwenaudio}.
In this work, we introduce the newest addition to the Qwen family of large language models and large multimodal modles: \textbf{Qwen2}.
Qwen2 is a series of LLMs, grounded in the Transformer architecture~\citep{transformer}, trained using next-token prediction.
The model series encompasses foundational, i.e., base language models, pre-trained but unaligned to human preferences, and instruction-tuned models, fine-tuned with single-turn and multi-turn instruction-following datasets suitable for chat and agent purposes.
Our release comprises four dense models with parameter counts of 0.5 billion, 1.5 billion, 7 billion, and 72 billion, plus a Mixture-of-Experts (MoE) model with 57 billion parameters, of which 14 billion are activated for each token.
The smaller models, specifically Qwen2-0.5B and Qwen2-1.5B, are designed for easy deployment on portable devices such as smartphones, earphones, and smart glasses. 
Conversely, the larger models cater to deployment across GPUs of varying scales.

All models were pre-trained on a high-quality, large-scale dataset comprising over 7 trillion tokens, covering a wide range of domains and languages. 
Compared to previous editions of Qwen, Qwen2 includes a broader spectrum of linguistic data, enhancing the quantity and quality of code and mathematics content. 
This enrichment is hypothesized to improve reasoning abilities of LLMs. 
Regarding post-training, all models underwent supervised fine-tuning and direct preference optimization (DPO, \citealp{rafailov2024direct}), aligning them with human preferences through learning from human feedback. 
This process endows the models with the capability to follow instructions effectively.

We have conducted a thorough evaluation of Qwen2, alongside a selection of baseline models including both open-weight and proprietary models accessible via API.
Qwen2 outperforms competing models in evaluations of both fundamental language capabilities and instruction-tuned functionalities
Specifically, Qwen2-72B-Instruct, our instruction-tuned variant, scores 9.1 on MT-Bench~\citep{mtbench}, 48.1 on Arena-Hard~\citep{arena}, and 35.7 on LiveCodeBench~\citep{livecodebench}.
Meanwhile, Qwen2-72B, the base language model, achieves 84.2 on MMLU~\citep{mmlu}, 37.9 on GPQA~\citep{gpqa}, 64.6 on HumanEval~\citep{humaneval}, 89.5 on GSM8K~\citep{gsm8k}, and 82.4 on BBH~\citep{bbh}.

%% file: content/model_tokenizer.tex
\section{Tokenizer \& Model}

\label{sec:arch}

This section introduces the tokenizer and model design of Qwen2.
We detail the model architecture and configurations for different model sizes.

\subsection{Tokenizer}

Following Qwen~\citep{qwen}, we employ the identical tokenizer based on byte-level byte-pair encoding. 
Notably, this tokenizer exhibits high encoding efficiency, as evidenced by its better compression rate relative to alternatives, facilitating the multilingual capabilities of Qwen2.

Models of all sizes employ a common vocabulary consisting of 151,643 regular tokens and 3 control tokens. 
For more information, please refer to \citet{qwen}.
It should be noted that, owing to considerations in distributed training, the effective size for the embeddings is larger.

\subsection{Model Architecture}

The Qwen2 series fundamentally constitute large language models based on the Transformer architecture, featuring self-attention with causal masks~\citep{transformer}. 
Specifically, this series encompasses dense language models of 4 scales and a Mixture-of-Experts (MoE) model.
We introduce the specifics of the dense models before delving into the MoE model's distinctive attributes.

\subsubsection{Qwen2 Dense Model}

The architecture of the Qwen2 dense models comprises multiple Transformer layers, each equipped with causal attention mechanisms and feed-forward neural networks (FFNs). Key differences from Qwen are described below:

\paragraph{Grouped Query Attention} 

We adopt Grouped Query Attention (GQA, \citealp{gqa}) instead of conventional multi-head attention (MHA). GQA optimizes KV cache usage during inference, significantly enhancing throughput. 
Detailed KV head configurations for various model sizes are reported in Section~\ref{sec:config}.

\paragraph{Dual Chunk Attention with YARN} 

To expand the context window of Qwen2, we implement Dual Chunk Attention (DCA, \citealp{chunkllama}), which segments long sequences into chunks of manageable lengths.
If the input can be handled in a chunk, DCA produces the same result as the original attention. 
Otherwise, DCA facilitates effective capture of relative positional information between tokens within and across chunks, thereby improving long context performance.
Moreover, we also employ YARN~\citep{yarn} to rescale the attention weights for better length extrapolation.

Moreover, we follow Qwen with the usage of SwiGLU~\citep{glu} for activation, Rotary Positional Embeddings (RoPE, \citealp{rope}) for positional embedding, QKV bias~\citep{qkv_bias} for attention, RMSNorm~\citep{rmsnorm} and pre-normalization for training stability.

\subsubsection{Qwen2 Mixture-of-experts Model}

The architecture of Qwen2 MoE models closely mirrors that of Qwen1.5-MoE-A2.7B~\citep{qwen_moe}. 
As a substitute for the original FFN, the MoE FFN consists of $n$ individual FFNs, each serving as an expert. 
Each token is directed to a specific expert $E_i$ for computation based on probabilities assigned by a gated network $G$:
\begin{align}
    \mathbf{p} &= \mathrm{softmax}\left(G\left(\mathbf{x}\right)\right), \\
    \mathbf{y} &= \sum\nolimits_{i \in \text{top}_k\left({\textbf{p}}\right)} \mathbf{p}_i E_i(\mathbf{x}).
\end{align}
In the following, we present critical design considerations of Qwen2 MoE.

\begin{table}[t]
\centering
\caption{\textbf{Architecture of Qwen2 dense and MoE models.} For MoE models, 57B-A14B denotes that the model has 57B parameters in total and for each token 14B parameters are active, the Intermediate size denotes that of each expert, and \# Activated Experts excludes the shared experts.}
\label{tab:to_sota}
\begin{tabular}{@{}lccccc@{}}
\toprule
 \bf Configuration &\textbf{0.5B}  & \textbf{1.5B} & \textbf{7B}  &\textbf{72B} &\textbf{57B-A14B}\\ 
\midrule
Hidden Size & 896 & 1,536 & 3,584  & 8,192 & 3,584 \\
\# Layers & 24 & 28 & 28  & 80 & 28 \\
\# Query Heads & 14 & 12 & 28  & 64 & 28 \\
\# KV Heads  & 2 & 2 & 4 & 8  & 4 \\
Head Size & 64 & 128 & 128 & 128 & 128 \\
Intermediate Size & 4,864 & 8,960 & 18,944 & 29,568 & 2,560\\
\# Routed Experts & - & -&- & -& 64\\
\# Activated Experts & - & -&- &- & 8 \\
\# Shared Experts & - & -& -& -& 8\\
Embedding Tying  & True &  True  & False & False & False \\
Vocabulary Size & 151,646 & 151,646 & 151,646 & 151,646 & 151,646 \\ 
\# Trained Tokens & 12T & 7T & 7T & 7T & 4.5T \\
  
\bottomrule
\end{tabular}
\end{table}

\paragraph{Expert Granularity}

The key structural difference between MoE models and dense models is that MoE layers incorporate multiple FFNs, each serving as an individual expert. 
Consequently, one straightforward strategy to transition from a dense architecture to an MoE architecture is to set the parameters of each expert equal to those of a single FFN from the original dense model. 
For example, transitioning from Mistral-7B~\citep{mistral} to Mixtral 8x7B~\citep{mixtral}, involves activating two of the eight experts at a time. 
Differently, our model employs fine-grained experts~\citep{deepseekmoe}, creating smaller-scale experts while activating a greater number of experts simultaneously. Given an equal total number of expert parameters and activated parameters, fine-grained experts offer a richer set of expert combinations. 
By leveraging these fine-grained experts, Qwen2 MoE facilitates more diverse and dynamic expert utilization, thereby enhancing overall performance and adaptability.

\paragraph{Expert Routing}

The design of expert routing mechanisms is crucial for enhancing the performance of MoE models. 
Recently, there has been a notable trend towards integrating both shared and routing-specific experts within MoE layers \citep{deepspeedmoe, deepseekmoe}. 
We adopt this approach, as it facilitates the application of shared experts across various tasks while reserving others for selective use in specific routing scenarios. 
The introduction of shared and specialized experts offers a more adaptable and efficient method for developing MoE routing mechanisms.

\paragraph{Expert Initialization}

We initialize the experts in a similar way to upcycling~\citep{upcycle}, leveraging the weights of a dense model. 
In contrast, our approach emphasizes diversification among fine-grained experts to enhance the model's representational breadth.
Given the designated expert intermediate size $h_{\text{E}}$, the number of experts $n$, and the original FFN intermediate size $h_{\text{FFN}}$, the FFN is replicated $\left\lceil \nicefrac{n \times h_{\text{E}}}{h_{\text{FFN}}} \right\rceil$ times.
This replication ensures compatibility with the specified number of experts while accommodating any arbitrary expert intermediate size.
To promote diversity within each FFN copy, parameters are shuffled along the intermediate dimension.
This guarantees that each fine-grained expert exhibits unique characteristics, even across different FFN copies.
Subsequently, these experts are extracted from the FFN copies, and the remaining dimensions are discarded. 
For each fine-grained expert, 50\% of its parameters are randomly reinitialized. 
This process introduces additional stochasticity into expert initialization, potentially enhancing the model's capacity for exploration during training.

\subsubsection{Model Configuration}
\label{sec:config}

In the following, we provide the key configuration and information for the Qwen2 series. 

The Qwen2 series consists of models of 5 sizes, which are Qwen2-0.5B, Qwen2-1.5B, Qwen2-7B, Qwen2-57B-A14B, and Qwen2-72B. 
Table~\ref{tab:to_sota} lists the hyper-parameters and important information, e.g., the number of pre-trained tokens. %
Particularly, Qwen2-57B-A14B is upscaled from Qwen2-7B.
Notably, Qwen2 models demonstrate a substantially lower Key-Value (KV) size per token relative to Qwen1.5 models. 
This characteristic translates into a reduced memory footprint, particularly advantageous in long-context inference tasks.

%% file: content/pretraining.tex
\section{Pre-training}

\label{sec:pre}

In the pre-training of Qwen2, our efforts were focused on refining the dataset and investigating methods to handle extended context lengths effectively.

\subsection{Pre-training Data}

The pre-training of the Qwen2 models involves the development of a new, large-scale, high-quality multilingual dataset. 
This dataset represents an improvement over the corpora used in previous Qwen and Qwen1.5 models~\citep{qwen,qwen1.5}, enhancing the scale, quality, and diversity of the pre-training data in several key areas:

\paragraph{Quality Enhancement}
The filtering algorithm has been refined with additional heuristic and model-based methods, including the use of the Qwen models to filter out low-quality data. 
Moreover, these models are utilized to synthesize high-quality pre-training data.

\paragraph{Data Expansion}
Compared to Qwen1.5~\citep{qwen1.5}, we have collected a significantly larger volume of high-quality code, mathematics, and multilingual data, enhancing the model's capabilities in respective areas. 
This new dataset supports approximately 30 languages, such as English, Chinese, Spanish, French, German, Arabic, Russian, Korean, Japanese, Thai, and Vietnamese.

\paragraph{Distribution Improvement}
To ensure the model learns the distribution akin to human-like learning, we conduct experiments on scaled-down models to optimize the mixing of data from various sources and domains.

Based on these enhancements, the pre-training data was expanded from 3 trillion tokens in Qwen1.5~\citep{qwen1.5} to 7 trillion tokens. 
An attempt to further relax the quality threshold resulted in a 12 trillion token dataset.
However, the model trained on this dataset did not show a significant performance improvement over the 7 trillion token model.
It is suspected that increasing the volume of data does not necessarily benefit model pre-training. 
Considering training costs, we opted to use the higher-quality 7 trillion token dataset for training larger models, leaving further exploration for future model iterations.

All Qwen2 dense models, excluding Qwen2-0.5B, were pre-trained on this large-scale dataset of over 7 trillion tokens. 
Qwen2-0.5B were pre-trained using the 12 trillion token dataset.
The MoE model received an additional 4.5 trillion tokens of pre-training, in line with the principle of upcycling. 
Similar to previous Qwen models, high-quality multi-task instruction data is integrated into the Qwen2 pre-training process to enhance in-context learning and instruction-following abilities.

\subsection{Long-context Training}

To enhance the long-context capability of Qwen2, we augmented the context length from 4,096 tokens to 32,768 tokens during the concluding phase of pre-training. 
This expansion was complemented by the introduction of a significantly increased volume of high-quality, lengthy data.
In conjunction with these enhancements, we modified the base frequency of RoPE from 10,000 to 1,000,000 to optimize performance in long-context scenarios~\citep{ropeabf}.

To fully leverage the model's length extrapolation potential, we adopted the YARN mechanism~\citep{yarn} and the Dual Chunk Attention mechanism~\citep{chunkllama}.
These strategies enable the model to process sequences of up to 131,072 tokens while maintaining high performance, as evidenced by minimal perplexity degradation in preliminary experiments.

%% file: content/posttraining.tex
\section{Post-training}

\label{sec:post}

Following extensive large-scale pre-training, we engage in a post-training phase for Qwen2. 
This process is pivotal in enhancing its proficiency across a broad spectrum of domains, including coding, mathematics, logical reasoning, instruction following, and multilingual comprehension. 
Moreover, it ensures that the generation from the models is in harmony with human values, making it helpful, honest, and harmless. 
Unlike traditional methods that heavily rely on extensive human supervision, our approach focuses on scalable alignment with minimal human annotation~\citep{cao2024towards}. 
Specifically, we investigate methods to acquire high-quality demonstration and preference data for Supervised Fine-Tuning (SFT) and Reinforcement Learning from Human Feedback (RLHF), aiming to minimize the need for human labeling while maximizing the quality and reliability of the data.

\subsection{Post-training Data}

The post-training data primarily consists of two components: demonstration data $\mathcal{D} = \{(x_i, y_i)\}$ and preference data $\mathcal{P} = \{(x_i, y_i^+, y_i^-)\}$, where $x_i$ represents the instruction, $y_i$ represents a satisfactory response, and $y_i^+$ and $y_i^-$ are two responses to $x_i$, with $y_i^+$ being the preferred choice over $y_i^-$. 
The set $\mathcal{D}$ is utilized in SFT, whereas $\mathcal{P}$ is employed in RLHF.

The construction of training data entails a two-step process: collaborative data annotation and automated data synthesis. 
First, we extract the data ontology from large-scale instruction corpora, leading to a broad and diverse set of high-quality instructions. 
These instructions are systematically enhanced to incorporate greater complexity. 
Through human annotation, we obtain the target response $y_i$ and their positive and negative counterparts $(y_i^+, y_i^-)$.
Subsequently, a variety of automated alignment strategies are employed to synthesize a substantial volume of artificially annotated data across the domains of code, mathematics, instruction-following, creation, role-playing, and safety.

\subsubsection{Collaborative Data Annotation}

\paragraph{Automatic Ontology Extraction} 
The process initiates with the application of InsTag~\citep{lu2023instag}, an open-set fine-grained tagger, to extract the underlying ontology from a large-scale instruction dataset. 
Subsequent manual refinement ensures the accuracy of the extracted ontology.

\paragraph{Instruction Selection} 
Each instruction, with tags annotated, is evaluated for tag diversity, semantic richness, complexity, and intent completeness. 
Based on these criteria, we select a set of representative instructions~\citep{dong2023abilities}.

\paragraph{Instruction Evolution} 
To enrich the instruction dataset, a self-evolution strategy~\citep{zhao2024tree} is employed, prompting the Qwen models to add constraints or requirements to existing instructions, thereby increasing their complexity and ensuring a diverse range of difficulty levels within the dataset. 

\paragraph{Human Annotation} 
Multiple responses to an instruction are obtained using diverse generation strategies and Qwen models of different scales. 
Annotators rank these responses based on their preferences, ensuring the best response meets established criteria, yielding both demonstration and preference data.

\subsubsection{Automated Data Synthesis}

Maintaining the quality of annotations for responses to instructions presents significant challenges on a large scale, particularly those that require expertise, experience, carefulness, or patience.
To address these challenges, we devised various automated alignment strategies to synthesize data at scale.

\paragraph{Rejection Sampling}
For mathematical or similar tasks with definitive final answers, rejection sampling~\citep{yuan2023scaling} is applied to improve the quality of solutions. 
Large language models (LLMs) are tasked to generate multiple responses, namely the reasoning paths, for each instruction.
Paths that result in accurate conclusions and are considered reasonable by the model are preserved, serving as demonstration data.
Preference data is generated by contrasting correct and incorrect paths.

\paragraph{Execution Feedback}
For coding tasks, LLMs are employed to generate solutions and associated test cases. 
The efficacy of these solutions is evaluated by compiling and executing them against the test cases, thereby creating demonstration and preference data. 
This methodology is also applicable to assessing instruction following~\citep{dong2024autoif}. 
For each instruction with constraints, e.g., length limit, the LLM is tasked to generate a Python verification function to ensure the response aligns with the instruction requirements.

\paragraph{Data Repurposing}
Creating skilled responses in literary writing tasks is challenging for annotators without specialized training. 
To tackle this problem, we aggregate high-quality literary works from the public domain and employ LLMs to develop instructions with varying levels of detail. 
These instructions, paired with the original works, serve as demonstration data. 
For example, to compile roleplay data with vivid and engaging responses, we source detailed character profiles from knowledge repositories such as Wikipedia and instruct LLMs to generate corresponding instructions and responses~\citep{lu2024large}. 
This process, similar to a reading comprehension task, ensures that the integrity of the character's profile is maintained.

\paragraph{Constitutional Feedback} 
Constitutional AI refers to the process of guiding LLMs to generate responses based on predefined sets of principles~\citep{bai2022constitutional}. 
To ensure adherence to guidelines such as safety and values, a constitution dataset was compiled. 
This dataset delineates principles to be followed and those to be avoided. 
It was used to instruct LLMs to produce responses that either are aligned with or deviated from these guidelines, serving as a reference for demonstration and preference data.

\subsection{Supervised Fine-tuning}

We have assembled an extensive instruction dataset featuring more than 500,000 examples that cover skills such as instruction following, coding, mathematics, logical reasoning, role-playing, multilingualism, and safety. 
Our model was fine-tuned for two epochs with a sequence length of 32,768 tokens. 
To optimize learning, the learning rate was gradually decreased from $7 \times 10^{-6}$ to $7 \times 10^{-7}$. 
To address overfitting, we applied a weight decay of 0.1 and gradients were clipped at a maximum value of 1.0.

\subsection{Reinforcement Learning from Human Feedback}

Our training regime for RLHF comprises two sequential stages: offline and online training. 
In the offline training stage, we use a pre-compiled preference dataset $\mathcal{P}$ to maximize the difference in likelihood between \(y_i^+\) and \(y_i^-\) with Direct Preference Optimization (DPO, \citealp{rafailov2024direct}). 
In the online training stage, the model iteratively refines its performance in real-time, leveraging  reward models for immediate feedback.  
Specifically, we sample multiple responses from the current policy model, and the reward model selects the most and the least preferred responses, forming preference pairs that are used for DPO in each episode. 
Moreover, we employ Online Merging Optimizer~\citep{lu2024online} to mitigate the alignment tax, i.e., the performance degradation associated with aligning model generation with human preferences.

%% file: content/experiments.tex
\section{Evaluation}
\label{sec:experiment}

To thoroughly assess the Qwen2 models, consisting of both base and instruction-tuned models, we implement a comprehensive evaluation protocol. 
This protocol examines a range of competencies, including general knowledge understanding, language comprehension, generation, coding, mathematics, reasoning, and additional areas of expertise. 
Specifically, base models are assessed using established benchmark datasets for large language models (LLMs), with responses elicited through few-shot prompting, unless specified otherwise. 
For instruction-tuned models, in addition to benchmark evaluations, we prioritize human preference assessments.

\subsection{Base Language Models}

In this section, we illustrate the evaluation of the base language models of the Qwen2 series. 
Specifically, we evaluate the models on benchmark datasets for knowledge and basic capabilities and apply multilingual benchmark datasets to evaluate their support of languages. As there are multiple model sizes, we compare them with the state-of-the-art (SOTA) models of similar or larger sizes. 

\subsubsection{Core Capabilities}

\begin{table}[t]
\centering
\caption{\textbf{Performance of the 70B+ models.} We compare Qwen2-72B with the baselines, including Mixtral-8x22B, Llama-3-70B, Qwen1.5-110B, and Qwen1.5-72B. For most datasets, Qwen2-72B demonstrates advantages over the baselines.}
\label{tab:main-72}
\setlength{\tabcolsep}{4.5pt}
\begin{tabular}{@{}lccccc@{}}
\toprule
\textbf{Datasets} &\textbf{Mixtral-8x22B}  & \textbf{Llama-3-70B} & \textbf{Qwen1.5-72B} & \textbf{Qwen1.5-110B} &\textbf{Qwen2-72B} \\
\midrule
\multicolumn{6}{c}{\textit{English}} \\
\midrule
MMLU & 77.8  & 79.5 & 77.5 & 80.4 & \textbf{84.2} \\
MMLU-Pro & 49.5  & 52.8 & 45.8 & 49.4 & \textbf{55.6} \\
GPQA & 34.3  & 36.3 & 36.3 & 35.9 & \textbf{37.9} \\
Theorem QA & 35.9  & 32.3 & 29.3 & 34.9 & \textbf{43.1} \\
BBH  & 78.9   & 81.0 & 65.5 & 74.8 & \textbf{82.4} \\
HellaSwag  & \textbf{88.7}  & 88.0 &  86.0 & 87.5 & 87.6 \\
Winogrande  & 85.0  & \textbf{85.3} &  83.0 & 83.5 &  85.1 \\
ARC-C  & \textbf{70.7}  & 68.8 & 65.9 & 69.6 &  68.9 \\
TruthfulQA  & 51.0  & 45.6 & \textbf{59.6} & 49.6 & 54.8 \\
\midrule
\multicolumn{6}{c}{\textit{Coding}} \\
\midrule
HumanEval & 46.3  & 48.2 & 46.3 & 54.3 & \textbf{64.6}  \\
MBPP & 71.7  & 70.4 & 66.9 & 70.9 & \textbf{76.9}  \\
EvalPlus & 54.1  & 54.8 & 52.9 & 57.7 & \textbf{65.4}  \\
MultiPL-E & 46.7  & 46.3 & 41.8 & 52.7 & \textbf{59.6}  \\
\midrule
\multicolumn{6}{c}{\textit{Mathematics}} \\
\midrule
GSM8K & 83.7   & 83.0 & 79.5 & 85.4 & \textbf{89.5} \\
MATH  & 41.7  & 42.5 & 34.1 & 49.6 & \textbf{51.1} \\
\midrule
\multicolumn{6}{c}{\textit{Chinese}} \\
\midrule
C-Eval   & 54.6    &  65.2 &  84.1 & 89.1 &  \textbf{91.0} \\
CMMLU   & 53.4  & 67.2 & 83.5 & 88.3 & \textbf{90.1} \\
\midrule
\multicolumn{6}{c}{\textit{Multilingual}} \\
\midrule
Exam   & 63.5 &   70.0    &  66.4 &  75.6 &  \textbf{76.6} \\
Understanding &   77.7    &  79.9 &  78.2 & 78.2 &  \textbf{80.7} \\
Mathematics &   62.9    &  67.1 &  61.7 & 64.4 &  \textbf{76.0} \\
Translation &   23.3    &  \textbf{38.0} &  35.6 & 36.2 &  37.8 \\
\bottomrule
\end{tabular}
\end{table}

\paragraph {Benchmarks and Evaluation Protocol} 
The common practice of evaluating the core capabilities of base language models is the implementation of benchmark dataset evaluation with few-shot or zero-shot prompting. The evaluation mainly focuses on the model performance of natural language understanding, general question answering, coding, mathematics, scientific knowledge, reasoning, etc. The datasets for evaluation include MMLU~\citep{mmlu} (5-shot), MMLU-Pro~\citep{mmlupro} (5-shot), GPQA~\citep{gpqa} (5shot), Theorem QA~\citep{theoremqa} (5-shot), BBH~\citep{bbh} (3-shot), HellaSwag~\citep{hellaswag} (10-shot), Winogrande~\citep{winogrande} (5-shot), TruthfulQA~\citep{truthfulqa} (0-shot), ARC-C~\citep{arc} (25-shot), HumanEval~\citep{humaneval} (0-shot), MBPP~\citep{mbpp} (0-shot), EvalPlus\citep{evalplus} (0-shot), MultiPL-E~\citep{multiple} (0-shot on Python, C++, Java, PHP, TypeScript, C\#, Bash, and JavaScript), GSM8K~\citep{gsm8k} (5-shot), MATH~\citep{math} (4-shot), C-Eval~\citep{ceval} (5-shot), and CMMLU~\citep{cmmlu} (5-shot). 
Multilingual datasets can be grouped into four categories: (a) Exam: M3Exam (5-shot, we only choose examples that require no image), IndoMMLU~\citep{koto-etal-2023-indommlu} (3-shot), ruMMLU~\citep{rummlu-mera} (5-shot), and translated MMLU~\citep{msift} (5-shot on Arabic, Spanish, French, Portuguese, German, Italian, Japanese, and Korean); (b) Understanding: BELEBELE~\citep{belebele} (5-shot), XCOPA~\citep{xcopa} (5-shot), XWinograd~\citep{xwinograd} (5-shot), XStoryCloze~\citep{xstory_cloze} (0-shot) and PAWS-X~\citep{paws-x} (5-shot); (c) Mathematics: MGSM~\citep{flores} (8-shot CoT); and (d) Translation: Flores-101~\citep{flores} (5-shot).

\begin{table}[t]
\centering
\caption{\textbf{Performance of the 30B+ dense models and 40B+ MoE models.} %
Qwen2-57B-A14B, an MoE model with a total of 57 billion parameters and 14 billion activated parameters, is designed to match the performance of 30 billion parameter dense models. 
This comparison includes dense model baselines: Yi-1.5-34B and Qwen1.5-32B, as well as MoE baselines: Mixtral-8x7B and Jamba. 
Results demonstrate that Qwen2-57B-A14B achieves competitive performance overall, with a notable superiority in coding and mathematics tasks.%
}
\label{tab:main-57a14}
\begin{tabular}{@{}lccccc@{}}
\toprule
\textbf{Datasets} &\textbf{Jamba}  & \textbf{Mixtral-8x7B} & \textbf{Yi-1.5-34B} & \textbf{Qwen1.5-32B} &\textbf{Qwen2-57B-A14B} \\
\midrule
Architecture & MoE & MoE & Dense & Dense & MoE \\
\# Act Params & 12B & 12B & 32B & 34B & 14B \\
\# Params & 52B & 47B & 32B & 34B & 57B   \\
\midrule
\multicolumn{6}{c}{\textit{English}} \\
\midrule
MMLU & 67.4 & 71.8 & \textbf{77.1} & 74.3 & 76.5 \\
MMLU-Pro & - & 41.0 & \textbf{48.3} & 44.0 & 43.0 \\
GPQA & - & 29.2 & - & 30.8 & \textbf{34.3} \\
Theorem QA & - & 23.2 & - & 28.8 & \textbf{33.5} \\
BBH  & 45.4 &  50.3  & \textbf{76.4} & 66.8 & 67.0 \\
HellaSwag  & \textbf{87.1} &  86.5  & 85.9 &  85.0 & 85.2 \\
Winogrande  & 82.5 &  81.9  & \textbf{84.9} &  81.5 &  79.5 \\
ARC-C  & 64.4 &  \textbf{66.0}  & 65.6 & 63.6 &  64.1 \\
TruthfulQA  & 46.4 &  51.1  & 53.9 & 57.4 &  \textbf{57.7} \\
\midrule
\multicolumn{6}{c}{\textit{Coding}} \\
\midrule
HumanEval & 29.3 & 37.2 & 46.3 & 43.3 & \textbf{53.0}  \\
MBPP & - & 63.9 & 65.5 & 64.2 & \textbf{71.9}  \\
EvalPlus & - & 46.4 & 51.9 & 50.4 & \textbf{57.2}  \\
MultiPL-E & - & 39.0 & 39.5 & 38.5 & \textbf{49.8}  \\
\midrule
\multicolumn{6}{c}{\textit{Mathematics}} \\
\midrule
GSM8K & 59.9 &  62.5  & \textbf{82.7} & 76.8 & 80.7 \\
MATH  & - &  30.8  & 41.7 & 36.1 & \textbf{43.0} \\
\midrule
\multicolumn{6}{c}{\textit{Chinese}} \\
\midrule
C-Eval   & - &   -    &  - &  83.5 &  \textbf{87.7} \\
CMMLU   & - &   -    & 84.8 & 82.3 & \textbf{88.5} \\
\midrule
\multicolumn{6}{c}{\textit{Multilingual}} \\
\midrule
Exam   & - &   56.1    &  58.3 &  61.6 &  \textbf{65.5} \\
Understanding & - &   70.7    &  73.9 &  76.5 &  \textbf{77.0} \\
Mathematics & - &   45.0    &  49.3 &  56.1 &  \textbf{62.3} \\
Translation & - &   29.8    &  30.0 &  33.5 &  \textbf{34.5} \\
\bottomrule
\end{tabular}
\end{table}

\paragraph{Qwen2-72B} 
In terms of the largest model of Qwen2, we compare Qwen2-72B with competitive baseline open-weight models, including Mixtral-8x22B~\citep{mixtral}, Llama-3-70B~\citep{llama3}, as well as Qwen1.5-72B~\citep{qwen1.5} and Qwen1.5-110B~\citep{qwen1.5_110}. 
The results are reported in Table~\ref{tab:main-72}.
Qwen2-72B outperforms Llama-3-70B in general \textbf{knowledge} understanding on both MMLU and MMLU-Pro, achieving accuracy improvements of 4.7 and 2.8, respectively.
In \textbf{scientific} assessments, Qwen2-72B demonstrates superiority over Llama-3-70B with enhancements of 1.6 and 9.8  on GPQA and Theorem QA.
Upon enrichment of \textbf{coding} data, Qwen2-72B exhibits a significant 18.3 and 10.0 percentage point advantage over Qwen1.5-72B in HumanEval and MBPP evaluations.
Enhanced \textbf{mathematics}-related data allows Qwen2-72B to outperform Qwen1.5-72B by 10.0 and 17.0 percentage points in the GSM8K and MATH benchmarks.
Qwen2-72B displays \textbf{reasoning} capabilities equivalent to Llama-3-70B, considering BBH, Winogrande, and ARC-C, attributable to its improved coding and mathematical data.
In assessing language understanding in \textbf{Chinese}, Qwen2-72B significantly outperforms Mixtral-8x22B and Llama-3-70B, and also outperforms Qwen1.5-72B.

\paragraph{Qwen2-57B-A14B} 
For the evaluation of the MoE model, Qwen2-57B-A14B is compared against baselines of similar sizes. 
These baselines include other MoE models, such as Mixtral-8x7B~\citep{mixtral} and Jamba~\citep{jamba}, and dense models, such as Yi-1.5-34B~\citep{yi} and Qwen1.5-32B~\citep{qwen1.5}, both of which have approximately 30 billion parameters. 
The results are shown in Table~\ref{tab:main-57a14}.
We anticipate that Qwen2-57B-A14B, which activates 14 billion parameters, will match the performance of a 30 billion parameter dense equivalent Qwen2 model.
Our evaluation reveals that Qwen2-57B-A14B performs comparably to Yi-1.5-34B in natural language understanding tasks. 
Moreover, it outperforms the baseline models in coding and mathematics tasks. 
Additionally, Qwen2-57B-A14B demonstrates robust Chinese language understanding capabilities, rivaling the larger Qwen2-72B model.
In essence, Qwen2-57B-A14B is an efficient model that, while activating only 14 billion parameters per forward pass, maintains the performance level of a 30 billion parameter dense model.

\begin{table}[t]
\centering
\caption{\textbf{Performance of the 7B+ models.} We compare Qwen2-7B with previously released state-of-the-art 7B+ models including Mixtral-7B, Gemma-7B, Llama-3-8B, and our previous Qwen1.5-7B. Qwen2-7B demonstrates significant advantages over the baselines in most of the evaluation datasets.}
\label{tab:main-7}
\begin{tabular}{@{}lccccc@{}}
\toprule
\textbf{Datasets} &\textbf{Mistral-7B}  & \textbf{Gemma-7B} & \textbf{Llama-3-8B} & \textbf{Qwen1.5-7B} &\textbf{Qwen2-7B} \\
\midrule
\multicolumn{6}{c}{\textit{English}} \\
\midrule
MMLU & 64.2 & 64.6 & 66.6 & 61.0 & \textbf{70.3} \\
MMLU-Pro & 30.9 & 33.7 & 35.4 & 29.9 & \textbf{40.0} \\
GPQA & 24.7 & 25.7 & 25.8 & 26.7 & \textbf{31.8} \\
Theorem QA & 19.2 & 21.5 & 22.1 & 14.2 & \textbf{31.1} \\
BBH  & 56.1 &  55.1  & 57.7 & 40.2 & \textbf{62.6} \\
HellaSwag  & \textbf{83.2} &  82.2  & 82.1 & 78.5 & 80.7 \\
Winogrande  & 78.4 &  \textbf{79.0}  & 77.4 &  71.3 &  77.0 \\
ARC-C  & 60.0 &  \textbf{61.1}  & 59.3 & 54.2 &  60.6 \\
TruthfulQA  & 42.2 &  44.8  & 44.0 & 51.1 &  \textbf{54.2} \\
\midrule
\multicolumn{6}{c}{\textit{Coding}} \\
\midrule
HumanEval & 29.3 & 37.2 & 33.5 & 36.0 & \textbf{51.2}  \\
MBPP & 51.1 & 50.6 & 53.9 & 51.6 & \textbf{65.9}  \\
Evalplus & 36.4 & 39.6 & 40.3 & 40.0 & \textbf{54.2}  \\
MultiPL-E & 29.4 & 29.7 & 22.6 & 28.1 & \textbf{46.3}  \\
\midrule
\multicolumn{6}{c}{\textit{Mathematics}} \\
\midrule
GSM8K & 52.2 &  46.4  & 56.0 & 62.5 & \textbf{79.9} \\
MATH  & 13.1 &  24.3  & 20.5 & 20.3 & \textbf{44.2} \\
\midrule
\multicolumn{6}{c}{\textit{Chinese}} \\
\midrule
C-Eval   & 47.4 &   43.6   &  49.5 &  74.1 &  \textbf{83.2} \\
CMMLU   & - &   -    & 50.8 & 73.1 & \textbf{83.9} \\
\midrule
\multicolumn{6}{c}{\textit{Multilingual}} \\
\midrule
Exam   & 47.1 &   42.7    &  52.3 &  47.7 &  \textbf{59.2} \\
Understanding & 63.3 &  58.3    &  68.6 &  67.6 &  \textbf{72.0} \\
Mathematics & 26.3 &   39.1    &  36.3 &  37.3 &  \textbf{57.5} \\
Translation & 23.3 &   31.2    &  \textbf{31.9} &  28.4 &  31.5 \\
\bottomrule
\end{tabular}
\vspace{-1\baselineskip}
\end{table}

\paragraph{Qwen2-7B}
The 7B model is widely utilized, as it enables the execution in 16-bit floating points on accelerators equipped with 16GB memory. 
Our focus is on comparing this model with other leading 7B models, including Llama-3-8B, which has recently demonstrated exceptional performance in the Chatbot Arena~\citep{arena}. 
This comparison also includes Mistral-7B-v0.2~\citep{mistral}, Gemma-7B~\citep{gemma}, and our predecessor, Qwen1.5-7B~\citep{qwen1.5}.
The results can be found in Table~\ref{tab:main-7}.
Qwen2-7B demonstrates superior performance across most datasets compared to other models, particularly excelling in coding tasks, mathematics, and Chinese language tasks. It also shows strong performance in multilingual understanding and exams. 
This indicates that Qwen2-7B has been optimized for a wide range of language and logic-based tasks, showcasing its versatility and advanced capabilities.

\begin{table}[t]
\centering
\caption{\textbf{Performance of the smaller models.} We compare our Qwen2-0.5B and Qwen2-1.5B with the previous SOTA small models including Phi-2, Gemma-2B and Qwen1.5-1.8B. Qwen2-0.5B with a much smaller model size achieves competitive performance, and Qwen2-1.5B significantly outperforms Qwen2-0.5B.}
\label{tab:main-small}
\begin{tabular}{@{}lccccc@{}}
\toprule
\textbf{Datasets} &\textbf{Phi-2}  & \textbf{Gemma-2B} & \textbf{Qwen1.5-1.8B} & \textbf{Qwen2-0.5B} &\textbf{Qwen2-1.5B} \\
\midrule
\# Non-Emb Params & 2.5B & 2.0B & 1.2B & 0.3B & 1.2B \\
\midrule
MMLU & 52.7 & 42.3 & 46.8 & 45.4 & \textbf{56.5} \\
MMLU-Pro & - & 15.9 & - & 14.7 & 21.8 \\
Theorem QA & - & - & - & 8.9 & \textbf{15.0} \\
BBH  & \textbf{43.4} &  35.2  & 24.2 & 28.4 & 37.2 \\
HellaSwag  & \textbf{73.1} &  71.4  & 61.4 &  49.3 & 66.6 \\
Winogrande  & \textbf{74.4} &  66.8  & 60.3 &  56.8 &  66.2 \\
ARC-C  & \textbf{61.1} &  48.5  & 37.9 & 31.5 &  43.9 \\
TruthfulQA  & 44.5 &  33.1  & 39.4 & 39.7 &  \textbf{45.9} \\
\midrule
HumanEval & \textbf{47.6} & 22.0 & 20.1 & 22.0 & 31.1 \\
MBPP & \textbf{55.0} & 29.2 & 18.0 & 22.0 & 37.4  \\
\midrule
GSM8K & 57.2 &  17.7  & 38.4 & 36.5 & \textbf{58.5} \\
MATH  & 3.5 &  11.8  & 10.1 & 10.7 & \textbf{21.7} \\
\midrule
C-Eval   & 23.4 &   28.0    &  59.7 &  58.2 &  \textbf{70.6} \\
CMMLU   & 24.2 &   -    & 57.8 & 55.1 & \textbf{70.3} \\
\bottomrule
\end{tabular}
\vspace{\baselineskip}
\end{table}

\paragraph{Qwen2-1.5B \& Qwen2-0.5B}
To evaluate the performance of our smaller models, specifically Qwen2-1.5B and Qwen2-0.5B, we compare them against established baselines: Phi-2~\citep{phi2}, Gemma-2B~\citep{gemma}, and Qwen1.5-1.8B~\citep{qwen1.5}.
The results are given in Table~\ref{tab:main-small}.
In language understanding, Qwen2-1.5B outperforms Phi-2, a model trained on textbook-like data. 
For coding tasks, Qwen2-0.5B matches the performance of Gemma-2B and Qwen1.5-1.8B, while Qwen2-1.5B surpasses these baselines, except for Phi-2.
Both Qwen2 models exhibit superior performance in mathematics compared to their competitors.
In terms of general reasoning, we find that Phi-2 generally outperforms all others, which to some extent reflects the significance of textbook data for reasoning capabilities. 
In TruthfulQA, Qwen2-1.5B performs the best, demonstrating that smaller models does not necessarily suffer from hallucination.
In Chinese language understanding, both Qwen2 models outperform all the others, a trend consistent with larger models in their respective comparisons.

In general, the Qwen2 series demonstrates superior performance against the baselines across different model sizes.
Notably, Qwen2-72B exhibits the highest performance among all Qwen2 models, underscoring the efficacy of model size scaling. %

\subsection{Instruction-tuned Model}

To critically evaluate instruction-tuned models, we implement a multifaceted approach. 
Assessments of foundational skills and human preferences are conducted using open datasets and benchmarks. 
Our detailed in-house examinations further probe model competencies in key areas. 
A particular focus is placed on assessing long context capability.
Safety measures include multilingual safety assessments and red teaming exercises. 
The following sections detail the evaluation methods and their outcomes.

\subsubsection{Open Benchmark Evaluation}

To comprehensively evaluate the quality of instruction-tuned models, we compile automatic and human evaluation to assess the capabilities and human preference. 
For the evaluation of basic capabilities, we apply similar datasets in the pre-trained model evaluation, which target on natural language understanding, coding, mathematics, and reasoning. 
Specifically, we evaluate on MMLU, MMLU-Pro, GPQA, and Theorem QA for language understanding and knowledge, HumanEval, MBPP, MultiPL-E, and LiveCodeBench v1~\citep{livecodebench} for coding, GSM8K and MATH for mathematics.
Additionally, we assess the performance of human preference alignment and instruction following by evaluating on benchmarks including MT-Bench~\citep{mtbench}, Arena-Hard~\citep{arena-hard}, AlignBench~\citep{alignbench}, MixEval~\citep{mixeval} whose results approximate those of Chatbot Arena, and IFEval~\citep{ifeval}\footnote{For simplicity, we report the results of the subset \textit{strict-prompt}.} for instruction following.

\begin{table}[t]
\centering
\caption{\textbf{Performance of 70B+ instruction-tuned models.} We compare Qwen2-72B-Instruct with Mixtral-8x22B-Instruct, Llama-3-70B-Instruct, Qwen1.5-72B-Chat, and Qwen1.5-110B-Chat. ``-Instruct'' or ``-Chat'' is omitted in the table. Qwen2-72B-Instruct demonstrates advantages in core capabilities, and superior performance in human preference alignment.}
\label{tab:70b_instruct_result}
\setlength{\tabcolsep}{2.6pt}
\begin{tabular}{@{}lccccc@{}}
\toprule
\textbf{Datasets} & \textbf{Mixtral-8x22B}  & \textbf{Llama-3-70B} & \textbf{Qwen1.5-72B} & \textbf{Qwen1.5-110B} &\textbf{Qwen2-72B} \\
\midrule
\multicolumn{6}{c}{\textit{English}} \\
\midrule
MMLU  & 74.0 & 82.0 & 75.6 & 76.5 & \textbf{82.3} \\
MMLU-Pro & 56.1 & 56.2  & 51.7 & 50.5  & \textbf{64.4} \\
GPQA & \textbf{49.7}  & 41.9 & 39.4 & 32.8 & 42.4 \\
Theorem QA & 40.8  & 42.5 & 28.8 & 18.8 & \textbf{44.4} \\
\midrule
\multicolumn{6}{c}{\textit{Coding}} \\
\midrule
HumanEval & 73.8  & 81.7 & 71.3 & 74.4 & \textbf{86.0}  \\
MBPP & 75.9  & \textbf{82.3} & 71.9 & 76.4 & 80.2  \\
MultiPL-E & 61.1  & 63.4 & 48.1 & 55.4 & \textbf{69.2}  \\
LiveCodeBench \textit{v1} & 21.8  & 29.3 & 17.9 & 25.3 & \textbf{35.7}  \\
\midrule
\multicolumn{6}{c}{\textit{Mathematics}} \\
\midrule
GSM8K & 89.1 & 93.0 & 82.7 & 84.5  & \textbf{93.2} \\
MATH  & 47.4  & 50.4 & 42.5 & 42.0 & \textbf{69.0} \\
\midrule
\multicolumn{6}{c}{\textit{Alignment}} \\
\midrule
MT-Bench   & 8.66    &  8.95 &  8.61 & 8.88 &  \textbf{9.12} \\
MixEval  & 82.3  & 84.0 & 84.1 & 85.7 & \textbf{86.7} \\
Arena-Hard  & 36.4  & 41.1 & 36.1 & 39.8 & \textbf{48.1} \\
IFEval \textit{strict-prompt} & 67.1  & 77.3 & 55.8 & 57.5 & \textbf{77.6} \\
AlignBench  & -  & 7.42 & 7.28 & 7.87 & \textbf{8.27} \\
\bottomrule
\end{tabular}
\end{table}

\paragraph{Qwen2-72B-Instruct}
We compare Qwen2-72B-Instruct against the instruction-tuned models including Mixtral-8x22B-Instruct, Llama-3-70B-Instruct, as well as Qwen1.5-72B-Chat. 
The results are presented in Table~\ref{tab:70b_instruct_result}.
It can be found that a strong base language model can help boost the downstream performance of the instruction-tuned model. 
Specifically, Qwen2-72B-Instruct outshines its peers in areas such as language understanding, coding, and mathematics, with the exception of GPQA and MBPP.
Regarding human preference alignment and instruction following, Qwen2-72B has significant advantages over the baselines.
We assume this achievement is attributed to both the high-quality pre-trained model and improvements in both data and training techniques for post-training.

\begin{table}[t]
\centering
\caption{\textbf{Performance of 30B+ dense and 40B+ MoE instruction-tuned models.} We compare Qwen2-57B-A14B-Instruct with the similar-size MoE model Mixtral-8x7B-Instruct, 30B dense models such as Yi-1.5-34B-Chat and Qwen1.5-32B-Chat. ``-Instruct'' or ``-Chat'' is omitted in the table. Qwen2-57B-A14B-Instruct is competitive with the recent SOTA 30B dense models, and significantly outcompetes the MoE baseline.}
\label{tab:moe_instruct_result}
\begin{tabular}{@{}lcccc@{}}
\toprule
\textbf{Datasets} & \textbf{Mixtral-8x7B}  & \textbf{Yi-1.5-34B} & \textbf{Qwen1.5-32B} & \textbf{Qwen2-57B-A14B} \\
\midrule
Architecture & MoE & Dense & Dense & MoE \\
\# Act Params & 12B & 32B & 34B & 14B \\
\# Params & 47B & 32B & 34B & 57B   \\
\midrule
\multicolumn{5}{c}{\textit{English}} \\
\midrule
MMLU  & 71.4 & \textbf{76.8} & 74.8 & 75.4 \\
MMLU-Pro & 43.3 & 52.3 & 46.4 & \textbf{52.8} \\
GPQA & - & - & 30.8 & \textbf{34.3} \\
Theorem QA & - & - & 30.9 & \textbf{33.1} \\
\midrule
\multicolumn{5}{c}{\textit{Coding}} \\
\midrule
HumanEval & 45.1 & 75.2 & 68.3 & \textbf{79.9}  \\
MBPP & 59.5 & \textbf{74.6} & 67.9 & 70.9 \\
MultiPL-E & - & - & 50.7 & \textbf{66.4}  \\
LiveCodeBench \textit{v1} & 12.3 & - & 15.2 & \textbf{25.5}  \\
\midrule
\multicolumn{5}{c}{\textit{Mathematics}} \\
\midrule
GSM8K & 65.7 & \textbf{90.2} & 83.6 & 85.3  \\
MATH  & 30.7 & \textbf{50.1} & 42.4 & 49.1 \\
\midrule
\multicolumn{5}{c}{\textit{Alignment}} \\
\midrule
MT-Bench   & 8.30 & 8.50 & 8.30 & \textbf{8.55} \\
MixEval  & 70.0 & 81.7 & 81.0 & \textbf{82.3} \\
IFEval \textit{strict-prompt} & - & - & 50.3 & \textbf{59.9} \\
AlignBench  & 5.70 & 7.20 & 7.19 & \textbf{7.36} \\
\bottomrule
\end{tabular}
\end{table}

\paragraph{Qwen2-57B-A14B-Instruct}
For medium-size models, we compare Qwen2-57B-A14B-Instruct with Mixtral-8x7B-Instruct, another MoE baseline, as well as the dense SOTA models with over 30 billion parameters, e.g., Yi-1.5-34B-Chat and Qwen1.5-32B-Chat. 
The results are provided in Table~\ref{tab:moe_instruct_result}.
Compared with Qwen1.5-32B-Chat, Qwen2-57B-A14B-Instruct reaches superior performance in almost all benchmarks, and compared with the 30B SOTA model Yi-1.5-34B-Chat, Qwen2-57B-A14B-Instruct has gained advantages in most evaluations except for those for mathematics. 
In terms of the evaluation for alignment, the advantages of Qwen2-57B-A14B-Instruct are notably evident.

\begin{table}[t]
\centering
\caption{\textbf{Performance of 7B+ instruction-tuned models.} We compare Qwen2-7B-Instruct with the recent SOTA models with 7-9 billion parameters, including Llama-3-8B-Instruct, Yi-1.5-9B-Chat, GLM-4-9B-Chat, and Qwen1.5-7B-Chat. ``-Instruct'' or ``-Chat'' is omitted in the table. Qwen2-7B-Instruct demonstrates competitive performance against Llama-3-8B-Instruct.}
\label{tab:7b_instruct_result}
\begin{tabular}{@{}lccccc@{}}
\toprule
\textbf{Datasets} & \textbf{Llama-3-8B}  & \textbf{Yi-1.5-9B} & \textbf{GLM-4-9B} & \textbf{Qwen1.5-7B} &\textbf{Qwen2-7B} \\
\midrule
\multicolumn{6}{c}{\textit{English}} \\
\midrule
MMLU  & 68.4 & 69.5 & \textbf{72.4} & 59.5 & 70.5 \\
MMLU-Pro & 41.0 & - & - & 29.1 & \textbf{44.1} \\
GPQA & 34.2 & - & - & 27.8 & \textbf{34.3} \\
Theorem QA & 23.0 & - & - & 14.1 & \textbf{25.3} \\
\midrule
\multicolumn{6}{c}{\textit{Coding}} \\
\midrule
HumanEval & 62.2 & 66.5 & 71.8 & 46.3 & \textbf{79.9}  \\
MBPP & \textbf{67.9} & - & - & 48.9 & 67.2  \\
MultiPL-E & 48.5 & - & - & 27.2 & \textbf{59.1}  \\
LiveCodeBench \textit{v1} & 17.3 & - & - & 6.0 & \textbf{26.6}  \\
\midrule
\multicolumn{6}{c}{\textit{Mathematics}} \\
\midrule
GSM8K & 79.6 & 84.8 & 79.6 & 60.3 & \textbf{85.7} \\
MATH  & 30.0 & 47.7 & 50.6 & 23.2 & \textbf{52.9} \\
\midrule
\multicolumn{6}{c}{\textit{Alignment}} \\
\midrule
MT-Bench   & 8.05 & 8.20 & 8.35 & 7.60 & \textbf{8.41} \\
MixEval  & 75.0 & 74.2 & - & 71.4 & \textbf{76.5} \\
IFEval \textit{strict-prompt} & \textbf{72.1} & - & 69.0 & 38.3 & 54.7 \\
AlignBench  & 6.20 & 6.90 & 7.01 & 6.20 & \textbf{7.21} \\
\bottomrule
\end{tabular}
\end{table}

\paragraph{Qwen2-7B-Instruct}
Within the spectrum of 7B to 9B models, we compare Qwen2-7B-Instruct with Llama-3-8B-Instruct, Yi-1.5-9B-Chat, GLM-4-9B-Chat, and Qwen1.5-7B-Chat. 
The results can be found in Table~\ref{tab:7b_instruct_result}.
Qwen2-7B-Instruct demonstrates substantial advancements compared to its predecessor, Qwen1.5-7B-Chat, across comprehensive evaluations, notably achieving higher scores in coding and mathematics-related tasks.
Compared with the recent SOTA model, Llama-3-8B-Instruct, Qwen2-7B-Instruct demonstrates competitive performance and specifically it achieves superior performance in coding. Nonetheless, in terms of instruction following, Qwen2-7B-Instruct greatly falls behind the competitor. 
To address this limitation, we plan to augment the 7B model's instruction-following ability by enhancing the quality of post-training data, ensuring a more robust understanding and execution of complex commands.

\begin{table}[t]
\centering
\caption{\textbf{Performance of smaller instruction-tuned models.} We compare both Qwen2-0.5B-Instruct and Qwen2-1.5B-Instruct with Qwen1.5-0.5B-Chat and Qwen2-1.8B-Chat. ``-Instruct'' or ``-Chat'' is omitted in the table. Compared with the similar-size baselines, Qwen2 significant surpasses the performance of Qwen1.5.}
\label{tab:small_instruct_result}
\begin{tabular}{@{}lcccc@{}}
\toprule
\textbf{Datasets} & \textbf{Qwen1.5-0.5B} & \textbf{Qwen2-0.5B} & \textbf{Qwen1.5-1.8B} & \textbf{Qwen2-1.5B} \\
\midrule
MMLU & 35.0 & \textbf{37.9} & 43.7 & \textbf{52.4} \\
HumanEval & 10.4 & \textbf{29.9} & 27.4 & \textbf{47.0} \\
MBPP & 14.5 & \textbf{37.8} & 28.6 & \textbf{51.9} \\
GSM8K & 11.3 & \textbf{40.1} & 35.3 & \textbf{61.6} \\
IFEval \textit{strict-prompt} & 14.6 & \textbf{20.0} & 16.8 & \textbf{29.0} \\
\bottomrule
\end{tabular}
\end{table}

\paragraph{Qwen2-1.5B-Instruct \& Qwen2-0.5B-Instruct}
In the context of smaller models, we compare Qwen2-0.5B-Instruct with Qwen1.5-0.5B-Chat, and Qwen2-1.5B-Instruct with Qwen1.5-1.8B-Chat.
Notably, the complexity of certain datasets designed for larger models exceeds the capabilities of these smaller models; thus, our analysis focuses on a selected subset.
As detailed in Table~\ref{tab:small_instruct_result}, the Qwen2 models demonstrate a marked advantage over their predecessors in both core capabilities and instruction-following tasks.
The achievement mainly attributes to the scaling of pre-training data. %
Consequently, our results affirm that data scaling remains an effective strategy for enhancing model performance, even in the domain of sub-billion parameter models.

\subsubsection{In-house Automatic Evaluation}

\begin{table}[tbp]
\centering
\small
\caption{\textbf{Performances of Qwen2-Instruct models on our in-house Chinese automatic evaluation benchmark.} Scores of Qwen2 models surpassing their comparable-sized Qwen1.5 counterparts are in bold. Qwen2-57B-A14B-Instruct is compared with Qwen1.5-32B-Chat.}
\label{tab:eval_v10_CN}
\setlength{\tabcolsep}{4pt}
\begin{tabular}{@{}lccccccc@{}}
\toprule
\textbf{Models} & \textbf{Knowledge} & \textbf{Exam} & \textbf{Comprehension} & \textbf{Coding} & \textbf{Math} & \textbf{Reasoning} & \textbf{Avg.} \\
\midrule
\multicolumn{8}{c}{\textit{Proprietary LLMs}} \\
\midrule
GPT-4o-2024-05-13 & 66.68  & 69.04  & 76.85 & 59.58 & 71.16 & 69.94 & 68.87 \\
Qwen-Max-0428 & 76.65  & 74.80  & 73.66 & 49.48 & 66.01 & 70.84 & 68.57 \\
\midrule
\multicolumn{8}{c}{\textit{Qwen1.5 Series}} \\
\midrule
Qwen1.5-0.5B-Chat & 28.55  & 36.99  & 29.70 & 3.82 & 13.10 & 25.47 & 22.94 \\
Qwen1.5-1.8B-Chat & 30.31  & 44.98  & 44.81 & 6.86 & 29.85 & 34.61 & 31.90 \\
Qwen1.5-4B-Chat & 33.67  & 47.17  & 50.44 & 14.05 & 36.20 & 39.98 & 36.92 \\
Qwen1.5-MoE-A2.7B-Chat & 52.76  & 60.49  & 52.84 & 19.34 & 38.45 & 43.07 & 44.49 \\
Qwen1.5-7B-Chat & 56.77  & 59.36  & 55.50 & 18.85 & 46.41 & 48.77 & 47.61 \\
Qwen1.5-14B-Chat & 63.35  & 66.13  & 60.06 & 28.19 & 54.80 & 50.20 & 53.79 \\
Qwen1.5-32B-Chat & 68.63  & 67.59  & 64.67 & 35.28 & 60.62 & 62.87 & 59.94 \\
Qwen1.5-72B-Chat & 71.52  & 70.04  & 66.70 & 38.22 & 63.09 & 61.30 & 61.81 \\
Qwen1.5-110B-Chat & 76.26  & 74.00  & 71.25 & 44.25 & 64.92 & 64.47 & 65.86 \\
\midrule
\multicolumn{8}{c}{\textit{Qwen2 Series}} \\
\midrule
Qwen2-0.5B-Instruct & 28.18  & \textbf{38.09}  & \textbf{35.90} & \textbf{9.40} & \textbf{21.20} & \textbf{25.61} & \textbf{26.40} \\
Qwen2-1.5B-Instruct & \textbf{35.46}  & \textbf{51.93}  & 44.70 & \textbf{14.05} & \textbf{34.58} & \textbf{35.94} & \textbf{36.11} \\
Qwen2-7B-Instruct & \textbf{61.54}  & \textbf{66.66}  & \textbf{59.63} & \textbf{34.74} & \textbf{60.99} & \textbf{58.22} & \textbf{56.96} \\
Qwen2-57B-A14B-Instruct & 64.15  & \textbf{73.67}  & \textbf{67.52} & \textbf{40.66} & \textbf{63.90} & 59.89 & \textbf{61.63} \\
Qwen2-72B-Instruct & \textbf{76.19}  & \textbf{75.65}  & \textbf{74.72} & \textbf{49.53} & \textbf{70.80} & \textbf{70.59} & \textbf{69.58} \\
\bottomrule
\end{tabular}
\end{table}

\begin{table}[tbp]
\centering
\small
\caption{\textbf{Performances of Qwen2-Instruct models on our in-house English automatic evaluation benchmark.} Scores of Qwen2 models surpassing their comparable-sized Qwen1.5 and Llama-3 counterparts are in bold. Qwen2-57B-A14B-Instruct is compared with Qwen1.5-32B-Chat.}
\label{tab:eval_v10_EN}
\begin{tabular}{@{}lccccc@{}}
\toprule
\textbf{Models} & \textbf{Knowledge} & \textbf{Comprehension} & \textbf{Coding} & \textbf{Math} & \textbf{Avg.} \\
\midrule
\multicolumn{6}{c}{\textit{Proprietary LLMs}} \\
\midrule
GPT-4o-2024-05-13 & 87.29 & 76.30 & 55.87 & 84.99 & 76.11 \\
Qwen-Max-0428 & 80.73 & 71.63 & 48.76 & 79.12 & 70.06 \\
\midrule
\multicolumn{6}{c}{\textit{Qwen1.5 Series}} \\
\midrule
Qwen1.5-0.5B-Chat  & 30.12 & 25.44 & 1.78 & 15.48 & 18.21 \\
Qwen1.5-1.8B-Chat & 40.37 & 41.87 & 4.99 & 29.71 & 29.23 \\
Qwen1.5-4B-Chat & 51.44 & 50.16 & 15.45 & 44.83 & 40.47 \\
Qwen1.5-MoE-A2.7B-Chat & 61.64 & 54.79 & 21.28 & 50.46 & 47.04 \\
Qwen1.5-7B-Chat & 64.86 & 58.61 & 20.79 & 54.24 & 49.62 \\
Qwen1.5-14B-Chat & 74.41 & 59.80 & 28.18 & 66.91 & 57.32 \\
Qwen1.5-32B-Chat & 76.38 & 64.70 & 37.39 & 73.04 & 62.88 \\
Qwen1.5-72B-Chat & 77.59 & 67.58 & 37.30 & 73.76 & 64.06 \\
Qwen1.5-110B-Chat & 78.29 & 70.17 & 44.12 & 78.87 & 67.86 \\
\midrule
\multicolumn{6}{c}{\textit{Llama-3 Series}} \\
\midrule
Llama-3-8B-Instruct  & 71.01 & 64.71 & 42.56 & 65.82 & 61.03 \\
Llama-3-70B-Instruct  & 83.06 & 76.31 & 57.18 & 79.70 & 74.06 \\
\midrule
\multicolumn{6}{c}{\textit{Qwen2 Series}} \\
\midrule
Qwen2-0.5B-Instruct & \textbf{43.19} & \textbf{29.57} & \textbf{6.95} & \textbf{31.52} & \textbf{27.81} \\
Qwen2-1.5B-Instruct & \textbf{56.03} & \textbf{45.08} & \textbf{17.61} & \textbf{50.44} & \textbf{42.29} \\
Qwen2-7B-Instruct & \textbf{73.75} & 63.09 & 36.41 & \textbf{75.67} & \textbf{62.23} \\
Qwen2-57B-A14B-Instruct & \textbf{76.80} & \textbf{67.92} & \textbf{42.37} & \textbf{77.04} & \textbf{66.03} \\
Qwen2-72B-Instruct & 83.00 & 73.58 & 53.03 & \textbf{82.15} & 72.94 \\
\bottomrule
\end{tabular}
\end{table}

Despite a number of open benchmark datasets for the evaluation, we believe that it is far from sufficient to fully comprehend the capabilities of LLMs. 
Specifically, we have made a series of in-house datasets that assess different capabilities of the models, e.g., knowledge understanding, text generation, coding, etc. 
The evaluation is in Chinese and English.
The results are gathered in Table~\ref{tab:eval_v10_CN} and Table~\ref{tab:eval_v10_EN}, respectively.

\paragraph{Chinese Evaluation} 

For the evaluations in Chinese, we focus on comparing the performance of Qwen2 models with the Qwen1.5 counterparts. 
For the small models, Qwen2-1.5B-Instruct generally outperforms Qwen1.5-1.8B-Chat in almost all the evaluations even with fewer parameters. 
In terms of the comparison of 7B models, the advantages of Qwen2 are more significant. 
Noteworthy is Qwen2-72B's superior performance to Qwen1.5-110B-Chat, despite the latter's greatly more parameters.
The MoE model displays superior performance across most domains relative to Qwen1.5-32B-Chat, excluding knowledge understanding.
This discrepancy may be attributed to a short of pre-training tokens.
In the near future, we are about to continue the pre-training of the MoE model to discover its scaling behaviors.

\paragraph{English Evaluation}

For English, we compare Qwen2 with both Qwen1.5 and Llama-3. 
Similarly, the small models of Qwen2 significantly outcompete the Qwen1.5 counterparts. 
However, in comparison with Llama-3-70B, Qwen2-72B-Instruct is falling behind by small margins especially in comprehension and coding. 
We assume both the amount of English tokens for pre-training and the quantity and diversity of data for post-training lead to the performance gap in English.

\subsubsection{Long Context Capabilities}

Three methods to evaluate long context capabilities are employed: the Needle in a Haystack (NIAH, \citealp{niah}), NeedleBench~\citep{opencompass}, and LV-Eval~\citep{yuan2024lveval}.

\paragraph{Needle in a Haystack}

\begin{figure}[tbp]
    \centering
    \includegraphics[width=\textwidth]{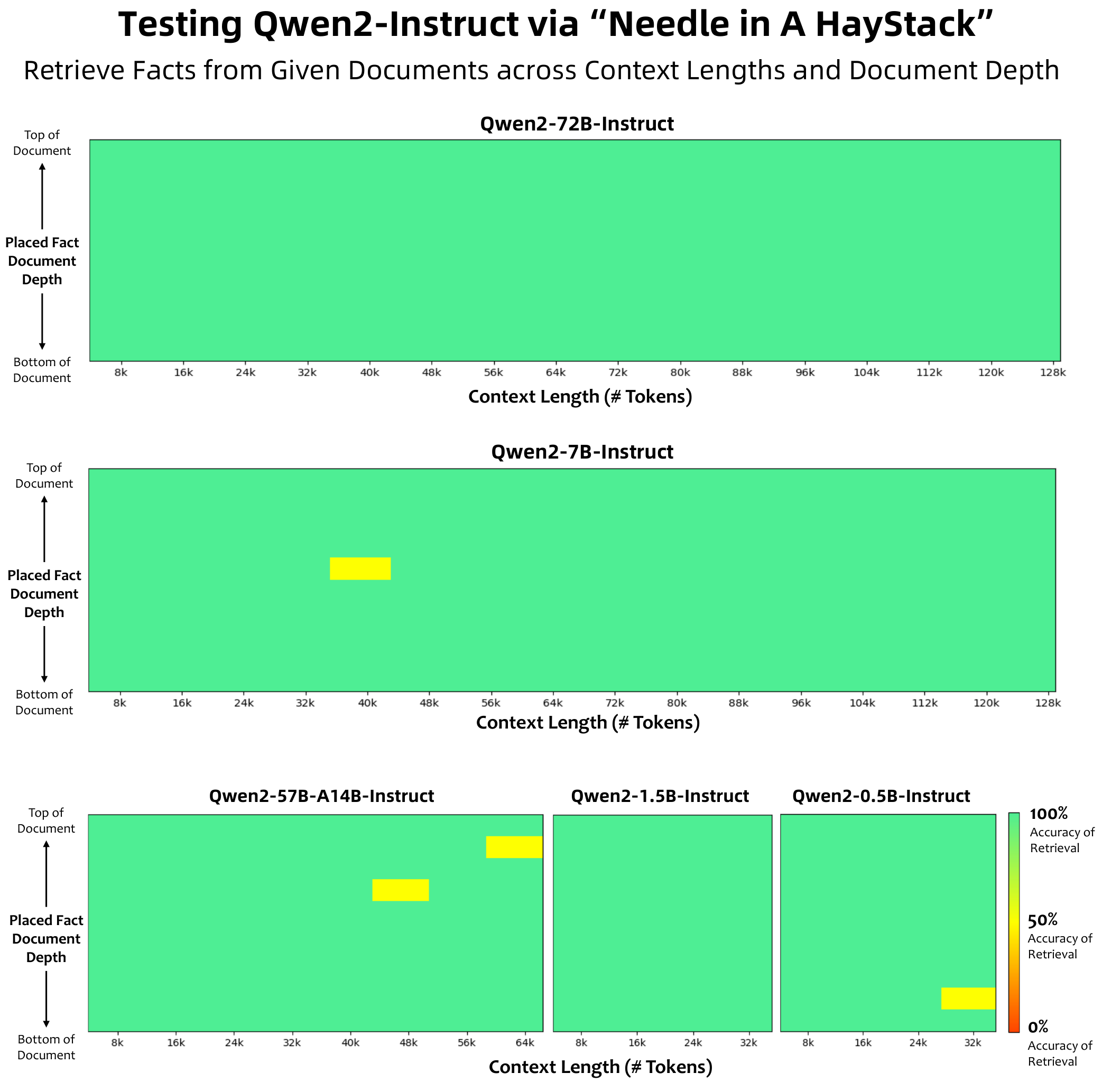}
    \caption{\textbf{Performance of Qwen2 instruction-tuned models on Needle in A Haystack Test.} All models that supports context lengths above 32k tokens integrates the YARN mechanism.}
    \label{fig:needle_in_haystack}
\end{figure}

This experiment assesses a model's proficiency in pinpointing facts within voluminous texts. 
Texts with 8K, 16K, ..., 128K tokens in length were crafted, with facts strategically positioned at varying depths. 
Each depth interval, e.g., from 0\% to 10\%, encompassed two instances. 
For contexts over 32K, YARN~\citep{yarn} was applied in this evaluation. 
As illustrated in Figure~\ref{fig:needle_in_haystack}, Qwen2-72B-Instruct exhibits exceptional accuracy in retrieving information from the entire 128K context. 
Coupled with its inherent strength, this model emerges as the optimal choice for processing extensive texts, assuming sufficient resources are accessible.
Additionally, models within the same series showcases remarkable performance across different context lengths. 
Precisely, Qwen2-7B-Instruct achieves a high level of accuracy in handling contexts up to 128K tokens. 
Meanwhile, Qwen2-57B-A14B-Instruct manages contexts up to 64K tokens proficiently, and the two smaller models in the Qwen2 series could support contexts of 32K tokens.

\begin{table}[tbp]
\centering
\caption{\textbf{Performance of Qwen2-72B-Instruct and Qwen2-7B-Instruct on NeedleBench and LV-Eval.} \textit{+YARN+DCA} does not change the model behavior within 32k tokens.}
\label{tab:longtext_benchmark}
\begin{tabular}{@{}lccccccccc@{}}
\toprule
\multirow{2}{*}{\textbf{Datasets}} & \multicolumn{4}{c}{\textbf{NeedleBench}} & \multicolumn{5}{c}{\textbf{LV-Eval}} \\\cmidrule(r){2-5}\cmidrule(l){6-10}
& \textbf{8k} & \textbf{32k} & \textbf{128k} & \textbf{256k} & \textbf{16k} & \textbf{32k} & \textbf{64k} & \textbf{128k} &  \textbf{256k} \\
\midrule
ChatGLM4-9B-1M & 56.61 & 49.15 & 44.30 & 45.29 & 46.40 & 43.23 & 42.92 & 40.41 & 36.95 \\
\midrule
Qwen2-7B-Instruct  & \multirow{2}{*}{87.07} & \multirow{2}{*}{73.64} & 38.77 & 2.92 & \multirow{2}{*}{49.77} & \multirow{2}{*}{46.93} & 28.03 & 11.01 & 0.55 \\
\ \ + YARN + DCA &  &  & 66.32 & 60.71 &  &  & 42.14 & 36.64 & 34.72  \\
\midrule
Qwen2-72B-Instruct  &  \multirow{2}{*}{\bf 91.90} & \multirow{2}{*}{\bf 92.01} & 73.05 & 17.13 & \multirow{2}{*}{\bf 58.82} & \multirow{2}{*}{\bf 56.70} & 42.92 & 31.79 & 2.88 \\
\ \ + YARN + DCA  &  &  & \bf 90.27 & \bf 85.21 & & & \bf 53.03 & \bf 48.83 & \bf \bf 42.35 \\
\bottomrule
\end{tabular}
\end{table}

\paragraph{NeedleBench}

NeedleBench ups the challenge on NIAH by including multiple facts (two to five) in passages, necessitating simultaneous identification and multi-hop reasoning. 
Table~\ref{tab:longtext_benchmark} reveals that the integration of YARN and DCA~\citep{chunkllama} notably improves Qwen2 models' long-context abilities. 
Qwen2-7B-Instruct surpasses ChatGLM4-9B-1M~\citep{chatglm4}, which claims a 1M context length.
Moreover, Qwen2-72B-Instruct demonstrates strong performance, with an accuracy reduction of just 6 points, compared to ChatGLM4-9B-1M, which shows a more pronounced decline of 11 points, particularly given its lower initial accuracy.

\paragraph{LV-Eval}

LV-Eval comprises 11 diverse QA datasets that demand comprehension of multiple pieces of evidence at once. 
To rectify the shortcomings of its original metric, which was excessively stringent and led to a high rate of false negatives, we adopt the keyword recall as the reported score. 
As shown in Table \ref{tab:longtext_benchmark}, integrating YARN and DCA substantially bolsters the long-context competencies of Qwen2 models on LV-Eval. 
Qwen2-7B-Instruct achieves parity with ChatGLM4-9B-1M, albeit with a more noticeable decline at extended contexts. Moreover, Qwen2-72B-Instruct demonstrates strong performance across all lengths, confirming its proficiency in handling long-context tasks.

\subsubsection{Multilingual Evaluation}

For the multilingual evaluation, we implement a comprehensive human evaluation for the assessment of multilingual capabilities. 
Specifically, we design diverse test cases assessing different capabilities of large language models, and we have test cases that are in a number of languages. 
For the annotators, we invite one professional annotator for each language who majors in the language for the evaluation. 
For each test case, the annotator grades the response from model with a score from 1 to 5.

\begin{table}[t]
\centering
\caption{\textbf{Performance of Qwen2-72B-Instruct and proprietary LLMs in multilingual human evaluation.} We compare Qwen2-72B-Instruct with GPT-3.5-Turbo-1106, GPT-4-Turbo-0409, GPT-4o-0513, Claude-3-Opus-0229. Scores range from 1 to 5. Overall, Qwen2-72B-Instruct performs substantially better than GPT-3.5-Turbo but there is progress to be made to be competitive with the proprietary models released in the last 6 months.}
\label{tab:multilingual_language}
\setlength{\tabcolsep}{4pt}
\begin{tabular}{@{}lccccc@{}}
\toprule
\textbf{Language} &\textbf{GPT-3.5-Turbo}  & \textbf{GPT-4-Turbo} & \textbf{GPT-4o} & \textbf{Claude-3-Opus} &\textbf{Qwen2-72B-Instruct} \\
\midrule
Arabic & 2.52 & 3.44 & 3.55 & 4.15 & 3.86 \\
French & 3.47 & 4.19 & 4.16 & 4.23 & 4.01 \\
Indonesian & 3.56 & 4.09 & 4.39 & 4.40 & 3.83 \\
Japanese & 2.75 & 3.68 & 3.72 & 3.85 & 3.63 \\
Korean & 2.37 & 4.24 & 4.40 & 4.23 & 4.14 \\
Portuguese & 3.37 & 3.86 & 3.89 & 4.09 & 3.97 \\
Russian & 3.24 & 4.27 & 4.32 & 4.25 & 4.15 \\
Spanish & 4.07 & 4.08 & 4.26 & 4.31 & 4.10 \\
Thai & 3.38 & 4.11 & 4.09 & 4.01 & 3.75 \\
Vietnamese & 3.90 & 3.84 & 4.14 & 3.98 & 3.91 \\
\midrule
Average & 3.16 & 3.98 & 4.09 & 4.15 & 3.93 \\
\bottomrule
\end{tabular}
\end{table}

We report the results of our model and the baselines in the evaluation of different languages. 
From Table~\ref{tab:multilingual_language}, it can be found that on average Qwen2-72B-Instruct significantly outperforms GPT-3.5-Turbo and it is competitive with GPT-4-Turbo and slightly falls behind Claude-3-Opus. 
This shows that our multilingual pre-training and instruction tuning data contribute to the multilingual capabilities of Qwen2-72B-Instruct and it is competitive with most state-of-the-art proprietary LLMs.

\subsubsection{Safety \& Responsibility}

\begin{table}[t]
\centering
\caption{\textbf{Performance of models in safety evaluation.} We compare Qwen2-72B-Instruct with GPT-4 and Mixtral-8x22B-Instruct. The lower, the better. Qwen2-72B-Instruct rejected more prompts with risks than the competitors.}
\label{tab:multilingual_safety}
\begin{tabular}{@{}lccc@{}}
\toprule
\textbf{Risk Category} &\textbf{GPT-4}  & \textbf{Mixtral-8x22B} & \textbf{Qwen2-72B-Instruct} \\
\midrule
Illegal & \phantom{0}0.00 & \phantom{0}6.87 & \phantom{0}0.00 \\
Fraud & \phantom{0}3.40 & \phantom{0}8.49 & \phantom{0}2.41 \\
Pornography & 23.63 & 33.82 & 22.91 \\
Privacy & \phantom{0}3.37 & 15.03 & \phantom{0}2.47 \\
\bottomrule
\end{tabular}
\end{table}

LLMs with openly accessible weights effectively accelerate the development of the research as well as their applications. 
Moreover, we believe that it is crucial to build safe and responsible LLMs so that the effect of the misuse of AI technologies could be significantly alleviated.

We implement a multilingual safety evaluation that tests the LLMs in different languages. 
Specifically, we assess the safety performance of the models in the topics about illegal behaviors, fraud, pornography, and privacy. 
We have collected prompts prone to jail-breaking and use them to test whether the models can provide safe responses by rejection. 

The results are presented in Table~\ref{tab:multilingual_safety}, where the proportion of harmful responses generated by the models are shown and the lower, the better.
It can be observed that Qwen2-72B-Instruct performs better than the proprietary model, GPT-4, and significantly outperforms the open-weight model, Mixtral-8x22B-Instruct. 
However, we believe that there is still much room for our model to improve to be a safer and more responsible model, especially in terms of pornography, which is a conventionally difficult category to differentiate even for humans.

\begin{table}[t]
\centering
\caption{\textbf{Contamination Analysis.} The \textit{contaminated} samples in this table are identified using a \textit{strict criterion}: any test sample with a 13-gram overlap with the pre-training or post-training data is considered contaminated. We report the percentage of contaminated samples as well as the model performance on both the original and non-contaminated test sets.}
\label{tab:contamination_analysis}
\setlength{\tabcolsep}{4pt}
\begin{tabular}{@{}lccccccc@{}}
\toprule
\multirow{2}{*}{\textbf{Test set}} & \textbf{Percent of}  & \multicolumn{3}{c}{\textbf{Qwen2-72B-Instruct}} & \multicolumn{3}{c}{\textbf{Qwen2-7B-Instruct}} \\
\cmidrule(r){3-5}\cmidrule(l){6-8}
& \textbf{\textit{Contamination}} & Original & Non-Contam. & $\Delta$ & Original & Non-Contam. & $\Delta$ \\
\midrule 
MMLU & 11.2\% & 82.3 & 83.2 & 0.9 & 70.5 & 71.3 & 0.8\\
MMLU-Pro & 11.6\% & 64.4 & 65.6 & 1.2 & 44.1 & 46.5 & 2.4\\
GPQA & \ \ 1.0\% & 42.4 & 41.8 & 0.6 & 34.3 & 34.1 & -0.2\\
HumanEval & 75.0\% & 86.0 & 87.0 & 1.0 & 79.9 & 87.8 & 7.9\\
MBPP & 29.6\% & 80.2 & 79.7 & 0.5 & 67.2 & 69.0 & 1.8\\
MultiPL-E & 37.7\% & 69.2 & 69.2 & 0.0 & 59.1 & 58.9 & -0.2\\
GSM8k & \ \ 0.7\% & 93.2 & 92.8 & -0.4 & 85.7 & 85.6 & -0.1\\
Math & 31.7\% & 69.0 & 74.6 & 5.6 & 52.9 & 57.6 & 4.7\\
IFEval & \ \ 0.9\% & 77.6 & 77.4 & -0.2 & 54.7 & 53.7 & -1.0\\
\bottomrule
\end{tabular}
\end{table}

\subsubsection{Contamination Analysis}

For large language models, what counts as contamination and how to run contamination analysis remain an active area of research~\citep{revaut2024how,golchin2024time,sainz2023nlp}. 
In the following, we first introduce how we try to decontaminate the training corpora against the evaluation datasets, and then estimate the extent to which benchmark scores are influenced by the remaining contamination.

During the construction of the pre-training and post-training datasets, we exclude potentially contaminated data using n-gram matching. 
However, we found that this approach may lead to a high false negative rate, because there could be commonly used expressions, especially in mathematical and coding data.
Therefore, we also applied another constraint based on the longest common subsequence (LCS).
Specifically, we first remove all symbols and punctuation from both the test and training sequences and perform tokenization. 
For a training sequence $\mathbf{s}_t$, we remove it if there is a test sequence $\mathbf{s}_e$ such that $|\text{LCS}(\mathbf{s}_t, \mathbf{s}_e)| \geq 13$ and $ |\text{LCS}(\mathbf{s}_t, \mathbf{s}_e)| \geq 0.6 \times \min(|\mathbf{s}_t|, |\mathbf{s}_e|)$.

To assess the potential effects of leaking data on the test performance, we follow \cite{gpt4} to construct a \textit{strict non-contaminated} test set to check if there is a significant performance degradation after \textit{strict decontamination}.
Specifically, we construct the non-contaminated test set by excluding any sample which has 13-gram overlap with the pre-training or the post-training data (without constraint on LCS), and then compute the corresponding metric on the test set.

The results are presented in Table~\ref{tab:contamination_analysis}. Although some datasets exhibit a high percentage of contamination under the strict criterion, we noticed that most of the identified \textit{contaminated} samples are false positives, primarily stemming from the mathematics and coding datasets. It is likely that certain code snippets and mathematical equations are so common that they do not provide any meaningful advantage in solving the test data. Furthermore, our analysis shows that the performance of the Qwen2 models remains consistent between the original and non-contaminated test data, suggesting that the potential issue of data contamination does not significantly impact the model's performance.

%% file: content/conclusion.tex
\section{Conclusion}
\label{sec:conclusion}

This technical report has presented the Qwen2 series, a versatile suite of foundational and instruction-tuned language models, ranging from 0.5 to 72 billion parameters, including models of dense and Mixture-of-Experts architecture. 
Qwen2 outperforms previous open-weight models, notably its predecessor Qwen1.5, and displays competitive performance against proprietary models across a broad spectrum of benchmarks in language understanding, generation, multilingual capabilities, coding, mathematics, and reasoning.
In this update, we have extra focus on long-context, multi-lingual, coding, mathematics capabilities and safety and responsibility.
In a commitment to fostering innovation and accessibility within the community, we have made the Qwen2 model weights openly accessible, which enables researchers and developers to harness the full potential of Qwen2 in a variety of applications and research projects. Through these efforts, we aim to contribute to the advancement of AI technologies and their positive impact on society.